\documentclass{article}

\usepackage{subcaption}
\usepackage{booktabs} 
\usepackage{float}

\usepackage{hyperref}       
\usepackage{url}            
\usepackage{amsfonts}       
\usepackage{nicefrac}       
\usepackage{microtype}      
\usepackage{amsmath}		
\usepackage{amssymb}
\usepackage[pdftex]{graphicx}	
\usepackage{caption}
\usepackage{cite}
\usepackage{bm}

\usepackage[accepted]{tccmlicml2021}

\renewcommand{\vec}[1]{\bm{\MakeLowercase{#1}}}
\newcommand{\mat}[1]{\bm{\MakeUppercase{#1}}}
\newcommand{\mean}[1]{\overline{#1}}
\newcommand{\pert}[1]{\delta #1}

\newcommand{\vk}{\vec{k}}
\newcommand{\vq}{\vec{q}}
\newcommand{\vt}{\vec{t}}
\newcommand{\vv}{\vec{v}}
\newcommand{\vw}{\vec{w}}

\newcommand{\vK}{\mat{K}}

\newcommand{\vT}{\mat{T}}
\newcommand{\vV}{\mat{V}}
\newcommand{\vW}{\mat{W}}

\newcommand{\vZ}{\mat{Z}}

\newcommand{\vpt}{\vec{\pert{t}}}

\icmltitlerunning{Ensemble Transformer in Neural Networks for Earth System Models}

\begin{document}
	
	\twocolumn[
		\icmltitle{Self-Attentive Ensemble Transformer: Representing Ensemble Interactions\\in Neural Networks for Earth System Models}
		
		
		
		\begin{icmlauthorlist}
			\icmlauthor{Tobias Sebastian Finn}{unihh,imprs}
		\end{icmlauthorlist}
		
		\icmlaffiliation{unihh}{Meteorological Institute, University of Hamburg, Hamburg, Germany}
		\icmlaffiliation{imprs}{International Max Planck Research School on Earth System Modelling, Max Planck Institute for Meteorology, Hamburg, Germany}
		
		\icmlcorrespondingauthor{Tobias Sebastian Finn}{tobias.sebastian.finn@uni-hamburg.de}
		
		\icmlkeywords{Machine Learning, Earth system modelling, Climate change, ensemble, transformer, self-attention}
		
		\vskip 0.3in
	]
	
	\printAffiliationsAndNotice{}

\begin{abstract}
	Ensemble data from Earth system models has to be calibrated and post-processed.
	I propose a novel member-by-member post-processing approach with neural networks.
	I bridge ideas from ensemble data assimilation with self-attention, resulting into the self-attentive ensemble transformer.
	Here, interactions between ensemble members are represented as additive and dynamic self-attentive part.
	As proof-of-concept, I regress global ECMWF ensemble forecasts to 2-metre-temperature fields from the ERA5 reanalysis.
	I demonstrate that the ensemble transformer can calibrate the ensemble spread and extract additional information from the ensemble.
	As it is a member-by-member approach, the ensemble transformer directly outputs multivariate and spatially-coherent ensemble members.
	Therefore, self-attention and the transformer technique can be a missing piece for a non-parametric post-processing of ensemble data with neural networks.
\end{abstract}

\section{Introduction}
In Earth system modelling, an ensemble of simulations \citep{leith_theoretical_1974} is a Monte-Carlo approach to estimate uncertainties in weather predictions \citep{bauer_quiet_2015,toth_ensemble_1993,molteni_ecmwf_1996} or to assess forced response and internal variability in the Earth system \citep{deser_insights_2020, kay_community_2015,maher_max_2019}.
Every ensemble members is physically-consistent in their multivariate structure.
The ensemble can thus naturally represent non-linear evolutions and non-Gaussian distributed states as they appear in nature.
Nevertheless, weather and climate ensembles have to be post-processed \citep{hemri_trends_2014,steininger_deep_2020} by model output statistics to correct model biases, calibrate the ensemble, and predict variables that are not modelled by the Earth system model.
Often, post-processing targets summarized ensemble statistics \citep{schulz_machine_2021}, predicting either the parameters \citep{gneiting_calibrated_2005,raftery_using_2005,rasp_neural_2018} or the cumulative distribution function \citep{taillardat_calibrated_2016,baran_combining_2018,bremnes_ensemble_2020,scheuerer_using_2020} of the target distribution.
As a consequence, the member-wise multivariate and spatial-coherent representation of the ensemble forecast is lost.
By contrast, I propose a member-by-member post-processing approach \citep{schaeybroeck_ensemble_2015} with neural networks and a self-attentive ensemble transformer that keeps the spatial correlation structure within the ensemble intact.

To calibrate the ensemble, ensemble members have to be informed about the evolution of other ensemble members.
The necessary term to represent the ensemble interactions is missing in neural networks that are applied on each ensemble member independently.
As a consequence, this direct neural network approach leads to a loss of information and to problems with tuning of the ensemble spread.

Ensemble Kalman filters \citep{evensen_sequential_1994,burgers_analysis_1998,bishop_adaptive_2001} include the dynamics between ensemble members by using the predicted ensemble covariances in a linear update step to assimilate given observations into ensemble predictions.
In their core approach, ensemble Kalman filters are similar to (self-)attention modules \citep{vaswani_attention_2017, luong_effective_2015,wang_non-local_2018} for neural networks, despite having another terminology: the value in attention modules or the state in ensemble Kalman filters is modified based on weights estimated with keys (the sensitivity in ensemble Kalman filters) and queries (observations).
In self-attention modules, the keys and queries are projections of the same input that is also used to project the values.
The module literally informs itself about the searched information.

\begin{figure*}[t]
	\centering
	\begin{subfigure}{.48\textwidth}
		\centering
		\includegraphics[height=0.2\textheight]{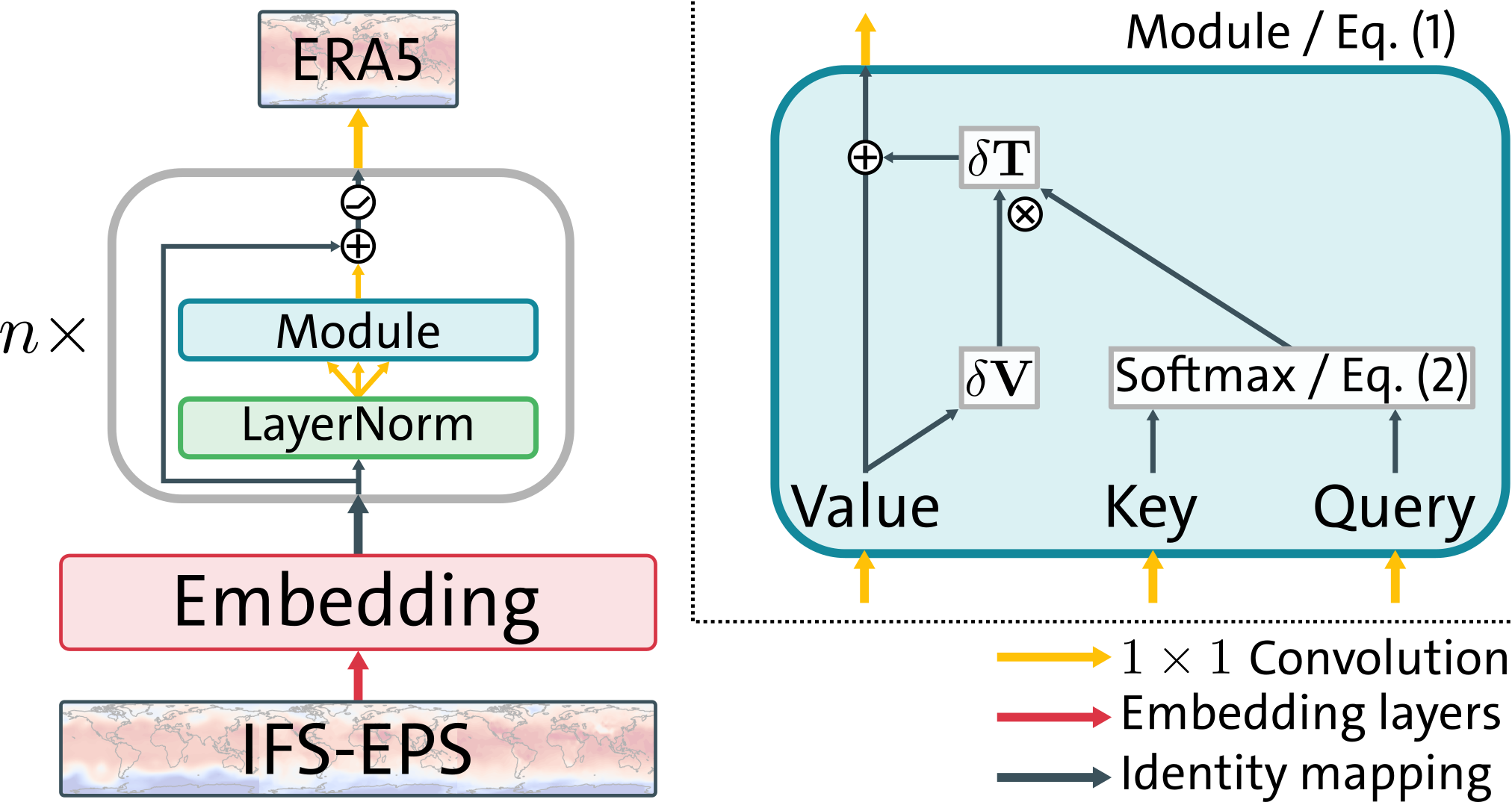}  
		\caption{
			Schematic overview of the ensemble transformer architecture.
			The separated side of the figure is a zoom-in to a single module.
		}
		\label{fig:schematic_overview}
	\end{subfigure}
	\hfill
	\begin{subfigure}{.48\textwidth}
		\centering
		\includegraphics[height=0.2\textheight]{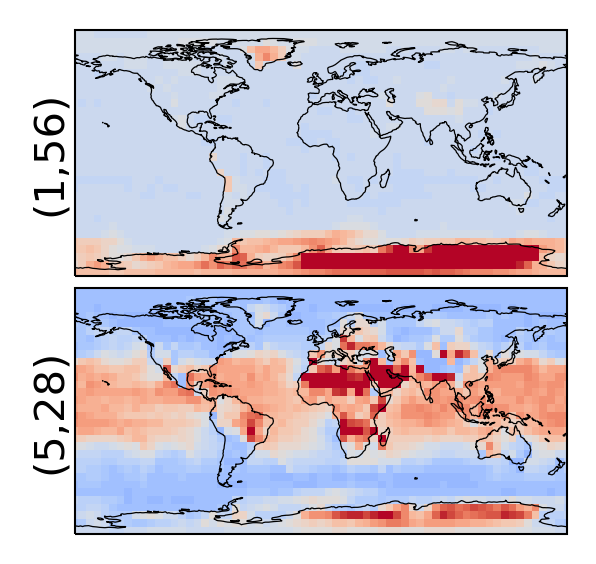}  
		\caption{Two selected attention maps (layer number, head number) from the Transformer (5) experiment for 2019-09-01 12:00 UTC. Red (blue) colours indicate a high (low) influence on the attention.}
		\label{fig:schematic_attention}
	\end{subfigure}
	\caption{
		Schematic overview of the self-attentive ensemble transformer architecture and two exemplary attention maps.\label{fig:schematic}
	}
\end{figure*}

I bridge the ideas of the ensemble Kalman filters and self-attention.
I introduce the self-attentive ensemble transformer for processing of ensemble data as neural network architecture by stacking multiple self-attention modules.
Each module adds to the static value for each ensemble member a dynamic self-attentive part that represents the interactions between ensemble members.
As these modules make use of the permutation-invariance of the ensemble members, this type of transformer can be seen as type of set transformer \citep{lee_set_2019}.
To test this idea and compare it to other methods, I regress global ECMWF ensemble forecasts to the 2-metre-temperature of the ERA5 reanalysis project as proof-of-concept experiments.

\section{The self-attentive ensemble transformer}\label{sec:transformer}

In the following, I introduce a single self-attentive transformer module as neural network layer.
A schematic overview over the architecture and module can be found in Figure \ref{fig:schematic_overview}.

Let ${\vZ_{l} \in \mathbb{R}^{k \times c \times h \times w}}$ be the input to the $l$-th layer with $k$ ensemble members, $c$ channels, $h$ latitudes, and $w$ longitudes.
The goal of the module is to estimate the transformed output $\vt_{i}(\vZ_{l}) \in \mathbb{R}^{c \times h \times w}$ of the $i$-th member based on the input of all members.
The transformed output is split into a static part $\vv_{i}$ and a dynamic part $\vpt_{i}(\vZ_{l})$.

The static part, also called value, encodes information that is only dependent on the current $i$-th ensemble member.
It is a linear projection of the input $\vV = \vZ_{l}\vW^{v}_{l}$ with a linear projection matrix $\vW^{v}_{l} \in \mathbb{R}^{c \times \tilde{c}}$ and $\tilde{c}$ number of channels in the attentive space, also called heads.

The dynamic part adds information from all members to the current $i$-th member.
I represent this as additive and linear combination of value perturbations with ensemble weights $\vw_{i} \in \mathbb{R}^{k \times \tilde{c}}$ and  $\mean{\vv} = k^{-1} \sum_{j=1}^{k} \vv_{j}$ as the ensemble mean of the values,
\begin{equation}
	\vt_{i}(\vZ_{l}) = \vv_{i} + \vpt_{i}(\vZ_{l}) = \vv_{i} + \sum_{j=1}^{k} \vw_{i,j}(\vv_{j}-\mean{\vv}).\label{eq:parametrization}
\end{equation}
In ensemble data assimilation, the update of ensemble predictions with observations is usually based on a similar parametrization \citep{bishop_adaptive_2001,lorenc_potential_2003,hunt_efficient_2007}.
Since no observations are available for post-processing purposes, the transformer module has to rely on self-attention.

In self-attention, the weights are estimated based on the same input data as the values \citep{vaswani_attention_2017,wang_non-local_2018}.
Here, the observations are replaced by a query $\vq_{i} \in \mathbb{R}^{\tilde{c} \times h \times w}$.
The query represents the searched information for the current $i$-th member and is estimated as linear projection of the input data with a projection matrix $\vW^{q}_{l} \in \mathbb{R}^{c \times \tilde{c}}$.
This query has to be related to the value perturbations of all members to estimate the weights.
The relation between query and values is established by a key matrix $\vK \in \mathbb{R}^{k \times \tilde{c} \times h \times w}$, which replaces the sensitivity matrix in data assimilation.
Again, a linear projection of the input data with a projection matrix $\vW^{k}_{l} \in \mathbb{R}^{c \times \tilde{c}}$ is used for the key matrix.

The weight are estimated based on the similarity between the query and key matrix.
In correspondence to \citet{vaswani_attention_2017}, the similarity is a scaled-dot product $\vK (\vq_{i})^T \in \mathbb{R}^{k \times \tilde{c}}$ over the latitudes and longitudes.
To obtain non-negative weights for a convex combination of value perturbations, the scaled-dot product is squashed through a softmax activation,
\begin{align}
	\vw_{i} &= \frac{\widetilde{\vw}_{i}}{\sum_{j=1}^{k}\widetilde{\vw}_{i,j}}, & \widetilde{\vw}_{i} = \exp(\frac{\vK (\vq_{i})^T}{\sqrt{h \times w}}).\label{eq:ens_self_attention}
\end{align}
These weights make thus explicitly use of the permutation-invariance in self-attention for ensemble data.

I model the output of the transformer module as residual connection \citep{he_deep_2015} with one residual branch and one identity mapping.
The residual branch is based on all transformed ensemble members $\vT(\vZ_{l}) \in \mathbb{R}^{k \times \tilde{c} \times h \times w}$, all estimated with \eqref{eq:parametrization} at the same time.
These transformed ensemble members are linearly projected by $\vW^{o}_{l} \in \mathbb{R}^{\tilde{c} \times c}$ from the attentive space back into the original feature space of the identity mapping.
I initialize $\vW^{o}_{l}$ as all-zero matrix; thus, only the identity mapping is used at the beginning of the training.
The output of the residual layer is activated with an activation function $f_{l}$, here the recitified linear unit (ReLU), and results into the input $\vZ_{l+1}$ of the next layer,
\begin{equation}
	\vZ_{l+1} = f_{l}(\vZ_{l} + \vT(\vZ_{l})\vW^{o}_{l}).
\end{equation}
This finishes the description of a single transformer module.

Since $\vpt_{i}(\vZ_{l})$ is a convex combination of value perturbations, one single-layered ensemble transformer module might be not expressive enough.
To extract more complex and non-linear interactions between ensemble members, it might be advantageous to stack multiple modules onto each other.

The ensemble space ($k = 50$) is normally much smaller than the spatial space (in my case $h \times w = 2048$).
Because the weights are estimated in this ensemble space, global self-attention is performed efficiently by \eqref{eq:parametrization} and \eqref{eq:ens_self_attention}.
The costs of the ensemble transformer scales quadratically with the number of members, but the weight formulation allows training with another number of members than used for inference as I show later.

The channels $\tilde{c}$ within the attentive space are similar to multiple heads in standard self-attention as the dot product is estimated over spatial dimensions.
The channels can thus represent different attentive regions.
To discover such regions with high influence, the element-wise product $\mean{\vk} \cdot \mean{\vq \vphantom{\vk}} \in \mathbb{R}^{\tilde{c} \times h \times w }$ of the ensemble mean key $\mean{\vk} = k^{-1} \sum_{j=1}^{k} \vk_{j}$ and the ensemble mean query $\mean{\vq} = k^{-1} \sum_{j=1}^{k} \vq_{j}$ can be used.
The here-exemplary shown maps (\mbox{Figure \ref{fig:schematic_attention}}) possibly represent regions with temperatures below the freezing level and with heat anomalies.

\section{Experiments and Discussion}\label{sec:experiments}

In a first step, I explain the used architectures and training methods.
As second step, I discuss and visualize the results from these experiments.

\subsection{Experimental strategy}

As input, I use data from the ECMWF ensemble prediction system (IFS-EPS, \citet{ecmwf_ifs_2019-1}) with $k=50$ ensemble members and three variables: the geopotential height on the $500\,\text{hPa}$ pressure level, the temperature on the $850\,\text{hPa}$ pressure level, and the 2-metre-temperature.
The forecasts with a lead time of 48 hours are valid for 00:00Z and 12:00Z.
They are fitted to the 2-metre-temperature of the ERA5 reanalysis project \citep{hersbach_era5_2020}.
The whole dataset consists of three-years data (2017-2019): 2017 and 2018 are used for training and validation, whereas 2019 is used for testing purpose.
I randomly select $10\,\%$ of 2017 and 2018 for validation.
As pre-processing, the global fields are bilinearly regridded to $h \times w = 32 \times 64$ grid points as in \citet{rasp_weatherbench_2020}.
The input data is normalized by their global mean and standard deviation, fitted for every variable independently based on the training dataset.

For all of my experiments, I use the same initial embedding structure with three consecutive two-dimensional convolutional layers, which are applied on every ensemble member independently.
For these convolutions, I use a kernel size of $5 \times 5$ with a locally-equidistant assumption, $c=64$ channels, and the ReLU activation.
I circularly pad in longitudinal direction and zero-pad in latitudinal direction.

In the \textbf{Transformer} experiments, I stack $n$ ensemble transformer modules between the embedding and the output.
For linear projections within the transformer layers, I use $1 \times 1$ convolutions with $\tilde{c}=64$ heads.
As proposed in \citep{xiong_layer_2020}, I apply layer normalization \citep{ba_layer_2016} across the channels, latitudes, and longitudes before the module input is linearly projected.
As output layer, I use a $1 \times 1$ convolution that combines the information from 64 channels into the 2-metre-temperature for each member independently.

As baseline, I perform to additional experiments with two other approaches.
First, I post-process each member independently with a neural network in the \textbf{Direct} experiments.
Secondly, I apply a parametric neural network (\textbf{PPNN}, \citet{rasp_neural_2018}) that outputs the mean and standard deviation as parameters of a Gaussian distribution.
In these parametric networks, the embedding output is averaged over all members and concatenated with the ensemble mean and standard deviation of the inputted 2-metre-temperature, similarly to \citet{rasp_neural_2018}.

In these baseline experiments, I replace the self-attention modules with $n$ residual layers \citep{he_deep_2015} between embedding and output.
These layers are two $1 \times 1$ convolutions with $64$ channels and the ReLU activation function in-between.
These residual layers have been modified with the fixed-update initialization in correspondence to \citep{zhang_fixup_2019}.
They are similar to the residual layer within the transformer module without self-attention.

As loss function, I minimize for all experiments the continuously ranked probability score (CRPS, \citet{hersbach_decomposition_2000, gneiting_strictly_2007}) with a Gaussian assumption and latitudinal weighting as in \citet{rasp_weatherbench_2020}.
For the transformer and direct experiments, I calculate the ensemble mean and the ensemble standard deviation from the resulting ensemble members as CRPS estimation step.
I have trained all models on a Nvidia GeForce GTX 1060 with a batch size of 8 samples.
Each experiment is optimized with Adam \citep{kingma_adam_2017} and an initial learning rate of $1\times10^{-3}$.
If the validation CRPS is not decreasing for 5 epochs, the learning rate is multiplied with $0.3$ of its previous value.
The training is ended if the validation CRPS is not decreasing for 20 epochs or after 200 epochs.
I have implemented\footnote[1]{Implementation can be found under: \url{https://github.com/tobifinn/ensemble_transformer}} the models with \hbox{PyTorch \citep{paszke_pytorch_2019}}.

\subsection{Results}

To compare the experiments (\hbox{Table \ref{tab:scores_subsampling}} and \hbox{Table \ref{tab:scores}}), I evaluate the latitudinal weighted spatio-temporal mean CRPS to the ERA5 reanalysis, the weighted spatio-temporal root-mean-squared-error of the ensemble mean (RMSE), and the square-root of the latitudinal weighted spatio-temporal mean of the ensemble variance (Spread).
If the ensemble spread is calibrated, it should match the RMSE.

\begin{table}[ht]
	\caption{
		The CRPS to the reanalysis, the ensemble mean RMSE, and the mean ensemble spread in the test dataset for all 50 ensemble members.
		The number behind the experiments indicates how many members were subsampled in each training sample. \label{tab:scores_subsampling}
	}
	\centering
	\begin{tabular}{l|ccc}
		Name (members) & CRPS & RMSE (K) & Spread (K)\\
		\hline
		Transformer (10) &  0.42 &  0.91 &    0.91 \\
		Transformer (20) &  0.42 &  0.92 &    0.90 \\
		Transformer (50) &  0.42 &  0.92 &    0.89 \\
	\end{tabular}
\end{table}

The training speed depends on the number of ensemble members that are used during the training.
To reduce the trainings costs, the ensemble can be subsampled by randomly selecting fewer members for each training sample (Table \ref{tab:scores_subsampling}). 
Because of additional noise, smaller subsampled sizes help to regularize the networks, but a too small subsampled ensemble can lead to an unstable training.
To strike a balance, I subsample 20 members in each training sample for all subsequent experiments.

\begin{table}[ht]
	\caption{
		The CRPS, the ensemble mean RMSE, and the ensemble standard deviation in the test dataset.
		The number behind the experiments indicates how many additional layers between embedding and output layer are used.
		Bold values represent the best performing methods.\label{tab:scores}
	}
	\centering
	\begin{tabular}{l|ccc}
		Name (layers) & CRPS & RMSE (K) & Spread (K)\\		
		\hline
		Climatology      &  2.60 &  6.12 &    6.05\\
		IFS-EPS raw      &  0.52 &  1.12 &    0.73\\
		\hline
		PPNN (0)         &  0.44 &  0.96 &    0.87\\
		PPNN (1)         &  0.43 &  0.95 &    0.87\\
		PPNN (5)         &  0.42 &  0.93 &    0.87\\
		Direct (1)       &  0.45 &  0.95 &    0.70\\
		Direct (5)       &  0.45 &  0.96 &    0.70\\
		Transformer (1)  &  0.42 &  0.91 &    \textbf{0.91}\\
		\textbf{Transformer (5)}  &  \textbf{0.41} &  \textbf{0.90} &    \textbf{0.90}
	\end{tabular}
\end{table}

The general performance of all methods is bounded by the available information from the input fields as can be seen in Table \ref{tab:scores}.
Nevertheless, the PPNN and Transformer approaches scale slightly with increasing network depth that leads to lower RMSE and CRPS values with increasing number of layers.
Since the RMSE of the Transformer experiments is reduced compared to the Direct and PPNN experiments, the self-attention mechanism can extract additional information from the interactions between ensemble members.
In addition, the experiments with the Transformer have a perfect spread-skill ratio (a probability integral transform histogram is shown in the Appendix, Figure \ref{fig:pit_hist}), whereas the ensembles in the Direct experiments are too small and underdispersive.
Therefore, the self-attention mechanism enables ensemble calibrations with neural networks and a member-by-member approach.
As a result, the ensemble transformer is the best performing method even compared to the parametric PPNN approach.

\begin{figure}[ht]
	\includegraphics[width=0.48\textwidth]{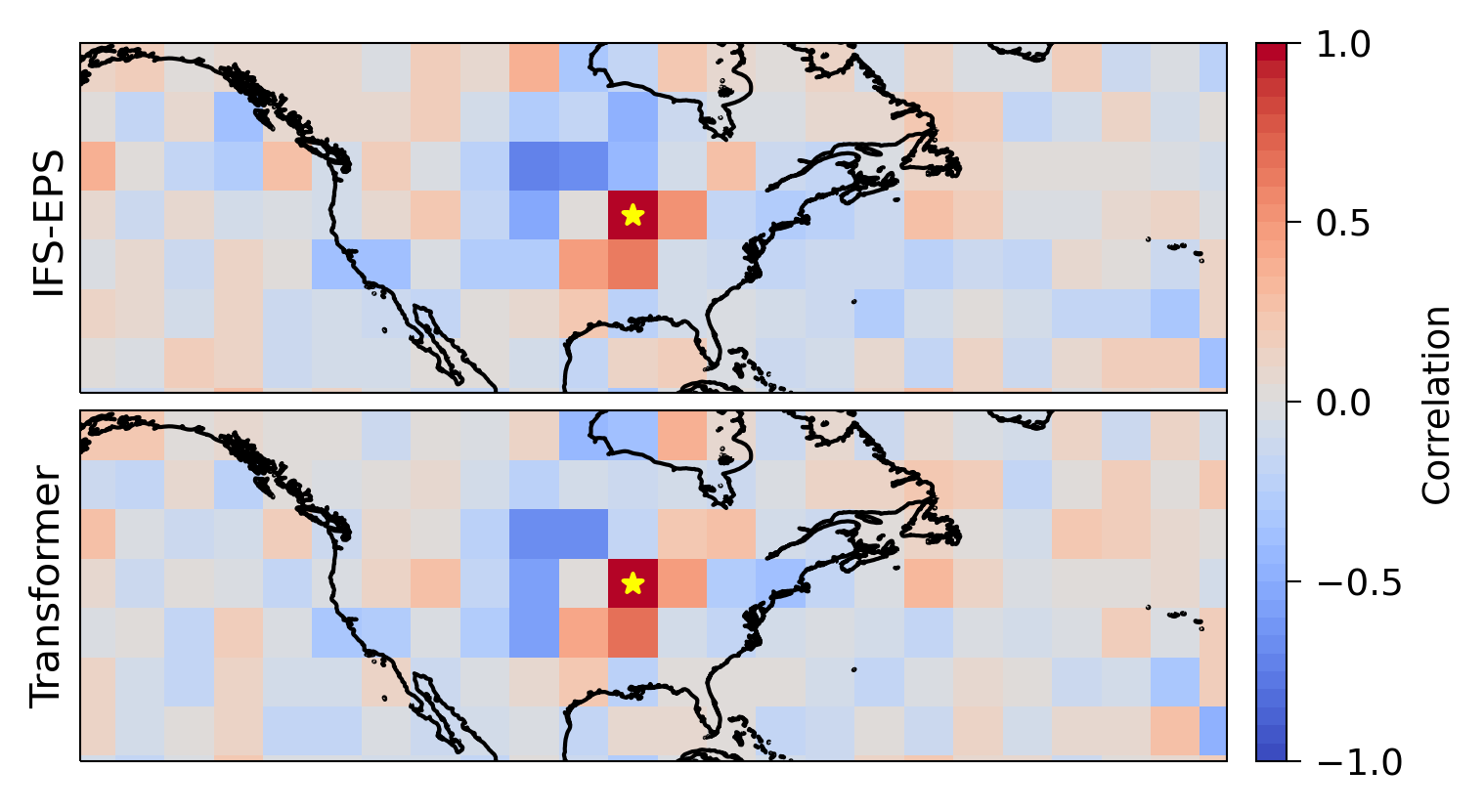}
	\caption{
		The spatial correlation within the 50 ensemble members from the IFS-EPS data and the Transformer(5) experiment at 26th January 2019, 12:00 UTC, estimated to the yellow-marked grid point of interest, roughly representing the position of Chicago.\label{fig:correlation}
	}
\end{figure}

Linear spatial patterns within the ensemble members can be found by analysing the ensemble correlation structure (Figure \ref{fig:correlation}).
Here, the post-processed ensemble members represent similar correlation structures as they can be found within the raw IFS-EPS ensemble.
Normally, additional methods like Gaussian copulas \citep{schefzik_uncertainty_2013,lerch_simulation-based_2020} are needed to represent such multivariate structures within a post-processed ensemble.
The ensemble transformer is a member-by-member approach and adds interactions between ensemble members as dynamic term.
It can thus directly output spatially-coherent ensemble members despite only targeting an univariately spatio-temporal averaged CRPS during training.

\section{Conclusion}
Based on the results of post-processing global ECMWF ensemble predictions to ERA5 2-metre-temperature reanalyses with ensemble transformers and convolutional neural networks, I conclude the following:
\begin{itemize}
	\item Self-attention can inform ensemble members about the evolution of other members within a neural network.
	Global self-attention can be hereby efficiently represented within the space of the ensemble members.
	\item The ensemble transformer can calibrate the ensemble spread.
	Furthermore, it can extract additional information from the interactions between ensemble members.
	\item Ensemble transformer can directly process ensemble members without using ensemble statistics and output again multivariate and spatially-coherent ensemble members.
\end{itemize}
Therefore, the self-attentive ensemble transformer can be a missing piece for a member-by-member post-processing of ensemble data with neural networks and without using summarized ensemble statistics.

Single model initial-condition large ensembles of climate simulations have to be calibrated \citep{suarez-gutierrez_exploiting_2021} for potential biases in the forced response and internal variability.
This study proofs that the training of self-attentive ensemble transformer for global post-processing of Earth system models is possible.
By leveraging historical runs and observations, such a transformer can be thus trained to calibrate these single model large ensemble.
This could then result in an improved assessment of the forced response and internal variability in the Earth system.

\section{Acknowledgements}
This work is a contribution to the research unit FOR2131, "Data Assimilation for Improved Characterization of Fluxes across Compartmental Interfaces", funded by the "Deutsche Forschungsgemeinschaft" (DFG, German Research Foundation) under grant 243358811.
I would like to acknowledge the ECMWF for providing the IFS-EPS data via the "The International Grand Global Ensemble" project and the Copernicus Climate Change Service (C3S) for distributing the ERA5 reanalysis data \citep{hersbach_era5_2020}, downloaded from the Climate Data Store.
I would like to thank Marc Bocquet, Sebastian Lerch, Laura Suarez-Gutierrez, and two anonymous reviewers for providing insightful remarks and suggestions that helped to improve the manuscript.

\small
\bibliography{icml_21}
\bibliographystyle{icml2021}

\normalfont
\appendix
\begin{minipage}[c]{0.48\textwidth}
	\section{Additional results}\label{sec:add_results}
	\includegraphics[width=\textwidth]{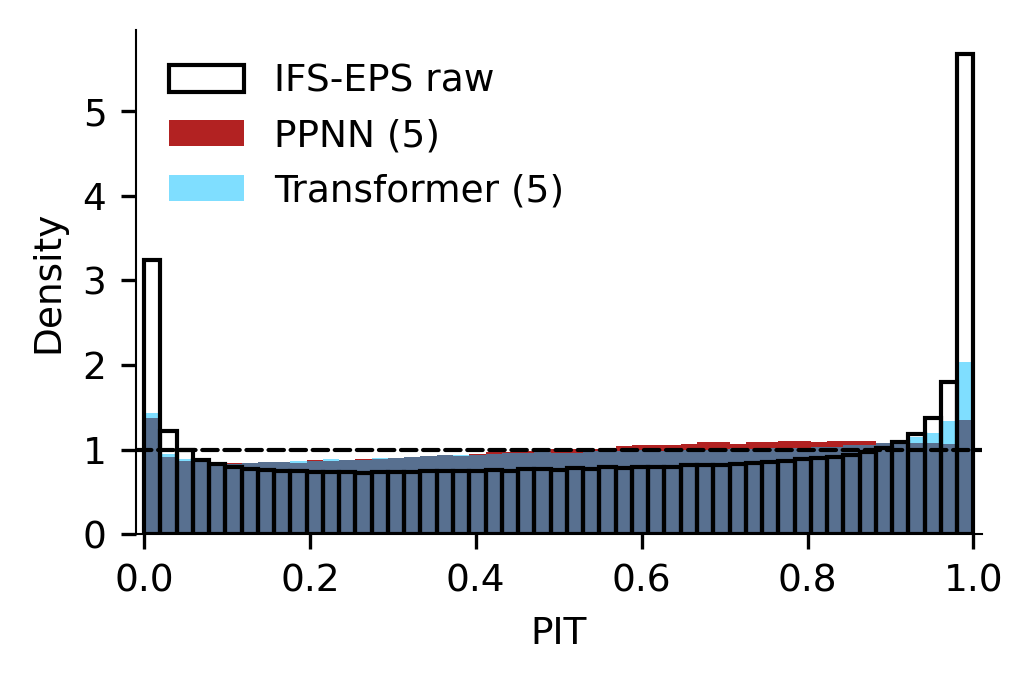}
	\captionof{figure}{
		Probability integral transform (PIT) histogram for the IFS raw data, the PPNN (5) experiment, and the Transformer (5) experiment for all grid point and time steps within the test dataset.
		The PIT histogram of the IFS-EPS raw and Transformer (5) experiment results out of a rank histogram.
		Because of the parametric approach, the PPNN histogram originates out of the Gaussian conditional probability functions.\label{fig:pit_hist}
	}
\end{minipage}

\end{document}